%% file: main.tex
\documentclass[sigconf, nonacm]{acmart}

\usepackage{enumitem}
\usepackage{graphicx}
\usepackage{microtype}
\usepackage{epstopdf}
\usepackage{flushend}
\usepackage{balance}
\usepackage{colortbl}
\usepackage[dvipsnames]{xcolor}
\usepackage{listings}
\usepackage{booktabs}
\usepackage{nicefrac}
\usepackage{caption}
\usepackage{subcaption}
\usepackage{soul}
\usepackage[normalem]{ulem} 
\usepackage{subfiles}
\usepackage{multirow}
\usepackage{enumitem}
\usepackage{tabularx}
\usepackage{xspace}
\usepackage{verbatim}
\usepackage{graphicx} 
\usepackage{pdfpages}

\usepackage[capitalize,nameinlink]{cleveref}
\crefformat{section}{\S#2#1#3}
\crefformat{subsection}{\S#2#1#3}

\newcommand{\sys}{\textsc{Credo}\xspace}

\newcommand{\abstractor}{abstractor\xspace}
\newcommand{\librarian}{librarian\xspace}
\newcommand{\compiler}{compiler\xspace}

\makeatletter
\newcommand{\CREDO@usbreak}{\textunderscore\allowbreak}
\DeclareTextFontCommand{\texttt}{\ttfamily
  \hyphenchar\font=`\-\relax      
  \language=\l@nohyphenation      
  \let\_\CREDO@usbreak}
\makeatother

\newcommand{\dto}{\allowbreak\ensuremath{\to}\allowbreak}
\newcommand{\dlr}{\allowbreak\ensuremath{\leftrightarrow}\allowbreak}

\newcommand{\dsl}{\,/\allowbreak\,}

\begin{document}

\title{Credo: Reusable Declarative Primitives for Agentic Workflows}

\author{Duo Lu}
\affiliation{%
  \institution{Brown University}
  \city{Providence}
  \state{RI}
  \country{USA}
}
\email{duo\_lu@brown.edu}

\author{Andrew Crotty}
\affiliation{%
  \institution{Northwestern University}
  \city{Evanston}
  \state{IL}
  \country{USA}
}
\email{andrew.crotty@northwestern.edu}

\author{U\u{g}ur \c{C}etintemel}
\affiliation{%
  \institution{Brown University}
  \city{Providence}
  \state{RI}
  \country{USA}
}
\email{ugur\_cetintemel@brown.edu}


\begin{abstract}
\input{0_abstract}
\end{abstract}

\maketitle

\input{1_introduction}
\input{2_credo}
\input{3_results}
\input{4_challenges}
\input{5_related}
\input{6_conclusion}

\bibliographystyle{ACM-Reference-Format}
\bibliography{references}

\end{document}

%% file: 0_abstract.tex
An LLM application depends on both a model and a \emph{harness}: the program that determines what each call sees, how many calls to make, and which answers to trust.
Coding agents can now discover strong harnesses by searching over candidate programs, but the resulting artifact is an opaque block of imperative code whose logical steps, runtime signals, physical execution decisions, and prompt strategies remain implicit and task-specific, forcing subsequent tasks to start the harness search process from scratch.

The potential for reuse, however, is substantial.
A searched harness encodes significant knowledge, such as the logical steps that work, the signals that matter, the physical operator decisions that adapt execution, and the prompt strategies that are effective, yet this knowledge is buried in imperative code with no inspectable or reusable structure, nor does it carry any provenance or metadata.

\sys addresses this problem by recovering a structured declarative description of a searched harness, tagging each extracted primitive with relevant metadata, and cataloguing all of it with provenance.
A compiler can then bind stored primitives to generate harnesses for new tasks without having to start the search over from scratch.
This paper provides preliminary results demonstrating the potential of our approach and lays out a related research agenda that the database community is well-positioned to tackle, including cost-based compilation over declarative catalogs and catalog maintenance under model and workload drift.

%% file: 1_introduction.tex
\section{Introduction}
\label{sec:intro}

\begin{figure*}[t]
  \centering
  \includegraphics[width=0.95\linewidth]{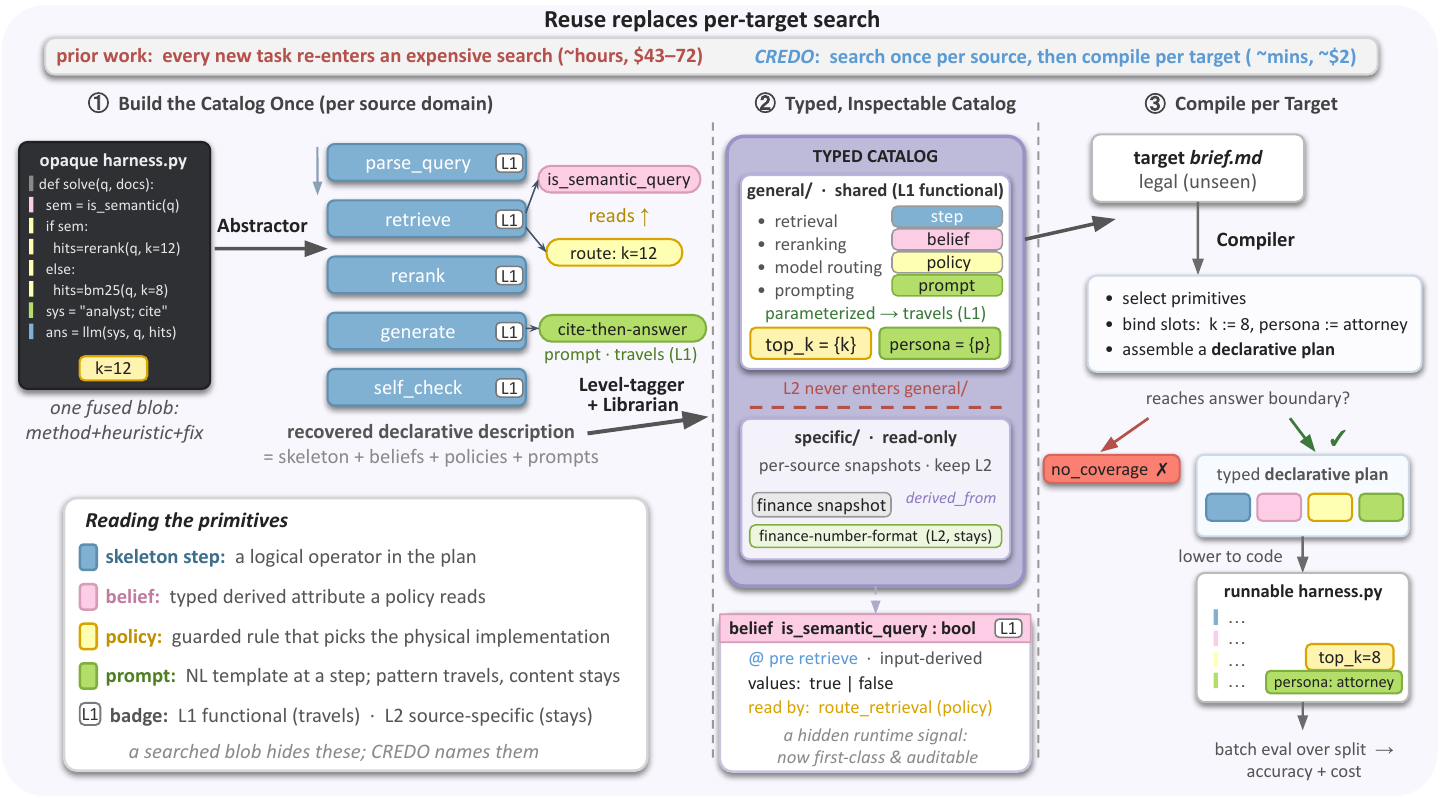}
  \caption{\sys recovers a declarative description from a
    searched harness once (\abstractor{}), files its transferable
    primitives in a shared catalog (level-tagger, \librarian{}), and
    later compiles a harness for an unseen target out of that catalog
    (\compiler{}) rather than searching again.}
  \label{fig:credo-overview}
\end{figure*}

The behavior of an LLM application depends heavily on its \emph{harness}, which determines the sequence of model and tool calls, the information supplied to each call, and the conditions under which an answer is accepted.
For example, a harness may begin with a small model, estimate confidence in the resulting answer, and invoke a larger model only when that estimate falls below a threshold.
Thus, both quality and cost depend on program-level choices that can extend well beyond the model itself.

Recently, agentic harness search has become an effective way to make these choices.
A search agent proposes candidate harnesses and executes them against a benchmark, using the observed accuracy and cost to guide subsequent iterations~\cite{metaharness}, following the evolutionary pattern coding agents apply to programs more generally~\cite{novikov2025alphaevolve}.
The final program encodes one successful combination of decisions as a single monolithic output, such that the logical sequence of operations, intermediate signals, conditional implementation choices, and prompt strategies all appear mixed together, along with constants and conventions specific to the benchmark.
As a result, knowledge that might be generalizable to other tasks is lost among the mess.

Database systems have faced a similar problem in the past.
For decades, user-defined functions (UDFs) remained as black boxes that could not be optimized or reasoned about, but work on UDF introspection and recovery of declarative expressions from procedural code has exposed opportunities that were previously hidden from the optimizer~\cite{ramachandra2017froid,duta2020plsql}.
We therefore argue for a similar approach for LLM harnesses, such that their internals can be decomposed into reusable declarative primitives.

\sys applies these lessons to LLM harnesses by attempting to recover a typed declarative description as primitives of four kinds: (1) \emph{skeleton steps} forming the logical plan; (2) \emph{beliefs} as derived attributes that inform physical operator selection; (3) \emph{policies} as the rules that select physical implementations using belief values; and (4) \emph{prompt templates} as the
parametric natural language specifications discovered at each step.
Every primitive is stored with provenance and relevant metadata, allowing a compiler to synthesize harnesses for new tasks from the catalog by reusing extracted primitives.
The key insight is that declarative descriptions of agentic execution are portable and durable, meaning they can be recovered once during harness search and then compiled into new harnesses across new domains, accumulating in a catalog that grows more valuable with every new task.
Our preliminary results demonstrate the promise of our proposed approach, which achieves more than an order of magnitude in cost savings over agentic harness search.
We also outline our vision for a broader research agenda this line of work opens up for the database community.

%% file: 2_credo.tex
\section{The \sys Framework}
\label{sec:framework}

The input to \sys is a searched \texttt{harness.py}, and its output is a declarative description from which a harness can be reconstructed or adapted.
The description must retain enough operational detail to reproduce behavior, while separating the method discovered by search from decisions tied to the source dataset.
\cref{fig:credo-overview} illustrates the process using a harness for a finance workload that is recovered into typed primitives, stored in a catalog, and subsequently recompiled into a harness for a legal workload.


\subsection{Structured Declarative Description}
\label{sec:description}

The declarative description encodes a harness as structured primitives expressed in natural language (NL).
Beliefs and policies keep the control-plane semantics of our earlier \sys system~\cite{lu2026credo}, where a developer authored them against their own pipeline, whereas here they are automatically extracted from a searched program, along with the skeleton and templates.
\sys extracts the following four types of reusable primitives.

\smallskip \noindent \textbf{Skeleton.}
The skeleton is the declarative logical plan, which is a directed acyclic graph (DAG) of named operator steps and the dependencies between them, recording what the harness computes and in what order.
The blue boxes in \cref{fig:credo-overview} are the skeleton the \abstractor{} recovers from the finance harness: \texttt{parse\_query}, \texttt{retrieve}, \texttt{rerank}, \texttt{generate}, and \texttt{self\_check}.
It is analogous to a query plan that describes the shape of the computation, separating logical intent (what each step does) from physical implementation (how it runs).
For now, we do not assume any formal equivalence rules for reordering or replacing steps, though this presents an interesting research opportunity (see \cref{sec:challenges}).

\smallskip \noindent \textbf{Beliefs.}
Beliefs are the derived attributes the harness computes at declared plan points and reads to make physical operator decisions.
Each has a name, a type from a closed value set \{bool, enum, int, float, str\}, a point in the skeleton where it becomes available, and the policies that consume it.
Beliefs fall into three categories: (1) \emph{input-derived} attributes computed from the raw input before any operator runs (e.g., \texttt{is\_semantic\_query~:~bool}); (2) \emph{intermediate-result} attributes computed from operator outputs and carried forward (e.g., \texttt{extraction\_confidence~:~float}); and (3) \emph{output-quality} attributes computed over final or near-final output to govern acceptance or escalation (e.g., \texttt{schema\_match~:~bool}).
In the finance harness, \texttt{is\_semantic\_query} was a local, implicit branch condition that \sys can promote to a declared, typed, named plan primitive so it can be logged, compared across runs, and traced to any associated policies that rely on it.

\smallskip \noindent \textbf{Policies.}
Policies are akin to physical operator selection rules in a database system.
Each policy evaluates a condition over one or more beliefs and, when it holds, selects the physical implementation of a logical step (e.g., model, parameters, execution strategy).
The yellow primitive in \cref{fig:credo-overview} is an example policy.
At \texttt{retrieve}, when \texttt{is\_semantic\_query} holds, select dense retrieval with \texttt{k=12}; otherwise, select keyword search with \texttt{k=8}.
Separating what a step is for from how it runs is the same as the logical/physical plan split in database query planning, where an operator's declarative behavior is decoupled from its actual implementation.

\smallskip \noindent \textbf{Prompt templates.}
Prompt templates are the parametric NL specifications the search discovered at each skeleton step (i.e., the physical operator implementations themselves).
Each template captures the reasoning strategy, instruction pattern, output format, or persona the search found effective, expressed as NL templates with bindable slots.
The two green primitives in \cref{fig:credo-overview} are templates, including a generic \texttt{cite-then-answer} pattern that travels to legal tasks and a finance-specific number-format nudge that stays local.

\begin{table*}[t]
  \caption{Evaluation datasets and the benchmarks each split
    draws from. The legal datasets are all taken from the
    LegalBench-RAG bundle~\cite{legalbench-rag}.}
  \label{tab:datasets}
  \begin{tabular}{@{}l l p{5.35cm} p{8.5cm}@{}}
    \toprule
    Dataset & Domain & Search / In-Distribution Sources &
    Cross-Distribution Sources \\
    \midrule
    Finance    & Retrieval
               & DocFinQA~\cite{docfinqa} + FinQA~\cite{finqa}
               & FinanceBench~\cite{financebench} + TAT-DQA~\cite{tat-dqa}
                 + TAT-HQA~\cite{tat-hqa} \\
    Legal      & Retrieval
               & CUAD-common~\cite{cuad} + MAUD~\cite{maud}
               & ContractNLI~\cite{contractnli} + PrivacyQA~\cite{privacyqa}
                 + CUAD-rare~\cite{cuad} \\
    Math       & Reasoning
               & OlympiadBench~\cite{olympiadbench} + Omni-MATH~\cite{omnimath}
               & IMO-AnswerBench~\cite{imo-answerbench-proofbench}
                 + IMO-ProofBench~\cite{imo-answerbench-proofbench}
                 + ArXivMath~\cite{arxivmath} \\
    Polyreason & Reasoning
               & GPQA-Diamond chem+bio~\cite{gpqa}
               & MMLU-Pro law+phil~\cite{mmlu-pro}
                 + ZebraLogicBench~\cite{zebralogic} + MuSR~\cite{musr} \\
    \bottomrule
  \end{tabular}
\end{table*}

\subsection{Building and Serving the Catalog}
\label{sec:library}

\sys processes the harness in two steps: (1) an offline catalog construction phase, and (2) an online compilation phase.
During catalog construction, the \abstractor{} extracts the primitives, the level-tagger determines their transfer scope, and the \librarian{} records the results.
Search and extraction are therefore performed for a source harness once, while later targets can reuse the resulting primitives through a substantially cheaper compilation step.

\smallskip \noindent \textbf{Building the catalog.}
Every primitive carries a transfer scope tag.
A primitive is \emph{functional} (L1) when its pattern recurs across a task family and can be lifted into the shared catalog; it is \emph{source-specific} (L2) when its pattern encodes one dataset's conventions and cannot be lifted out.
Scope is judged on the pattern, not the surface text or constants a primitive carries.
\sys first lifts source-specific constants into bindable slots (e.g., a retrieval step that hard-codes \mbox{\texttt{top\_k=12}} instead exposes \mbox{\texttt{top\_k=\{k\}}}), but parameterizing a constant does not by itself make a primitive transferable; rather, only a pattern that survives the lift does.
The tag is a binary gate, such that functional primitives enter the shared catalog and source-specific ones stay with their source.

Once a declarative description is recovered and its primitives are scope-tagged, the \librarian{} files them in a catalog with two tiers.
The \texttt{general/} tier is shared, holding the parameterized functional (L1) primitives indexed by function (e.g., retrieval, reasoning) rather than by subject, with each entry recording the provenance of which harness(es) it came from.
The \texttt{specific/} tier is the per-harness record, and it keeps the entire description as source-specific primitives, such that promoting part of a description to \texttt{general/} never discards the rest.
As the names suggest, the \texttt{general/} tier is primarily intended for reuse, whereas the \texttt{specific/} tier is only relevant for individual harnesses and primarily maintained for provenance purposes.
Thus, all of the resulting primitives remain both inspectable and traceable throughout.

\smallskip \noindent \textbf{Serving a target.}
The \compiler{} selects primitives to connect the target's input to an answer, binds their open slots to target-specific values, and assembles a runnable \texttt{harness.py}.
If no combination reaches a satisfactory threshold, \sys simply falls back to the harness search process.
Importantly, the catalog retains only primitive descriptions, not any implementation details.

These two phases also define how we evaluate \sys (\cref{sec:sem-fidelity} and \cref{sec:sem-composition}).
Recovering a declarative description and compiling it straight back, then running the original and the compiled harness on the same held-out split, is an empirical fidelity check, such that matching both accuracy and cost is evidence that the compiled harness faithfully captures the original's execution method.
Compiling a harness for a new target from primitives on other domains tests whether reuse can yield a working harness without a new search.
We discuss these further in the next section.

%% file: 3_results.tex
\begin{figure*}[t]
\centering
\includegraphics[width=\textwidth]{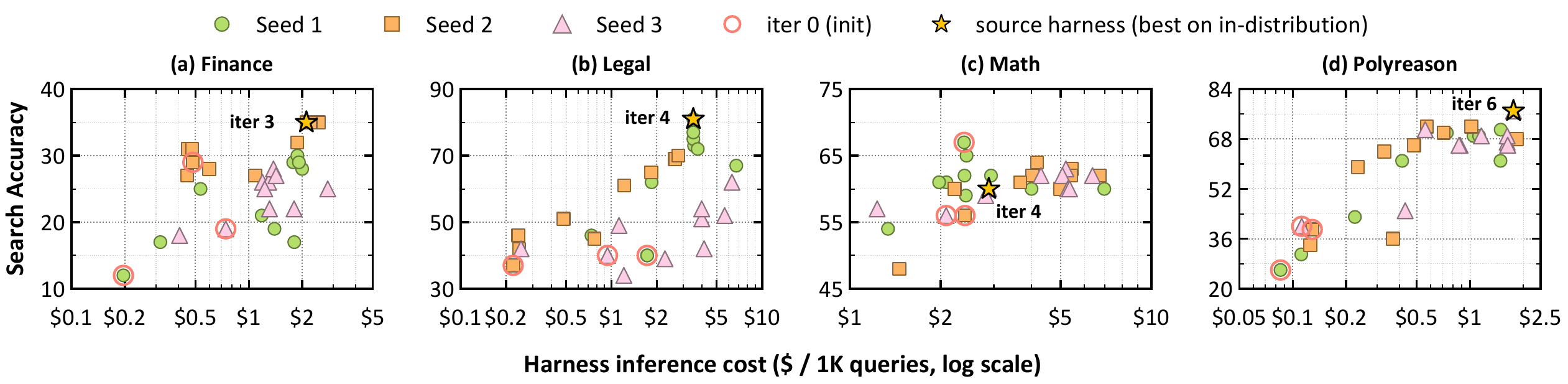}
\caption{Meta-Harness search trajectories, one plot per domain; each marker is one search iteration. Stars mark each domain's best program on the \emph{in-distribution} split, which need not be the topmost point on the search-split axis.}
\label{fig:mh-trajectory}
\end{figure*}

\section{Preliminary Results}
\label{sec:exp-setup}

This section describes our preliminary results.
We organize our experiments around the following research questions:
\begin{itemize}[wide=0pt]
  \item \textbf{Catalog Construction (\cref{sec:harness-discovery}).}
    Can \sys recover useful primitives from a searched harness?
  \item \textbf{Declarative Fidelity (\cref{sec:sem-fidelity}).} Can a recovered
    declarative description preserve the behavior of the harness it describes?
  \item \textbf{Primitive Reuse (\cref{sec:sem-composition}).} Can catalogued primitives be compiled into a working harness for an unseen target without re-searching?
\end{itemize}

\smallskip \noindent \textbf{System and Models.}
All experiments run on a single dual-socket server with 2$\times$ AMD EPYC~9554 processors and $755$\,GiB of RAM, hosting 4$\times$ NVIDIA L40S GPUs ($48$\,GB each, $192$\,GB aggregate).
The open-weight deployment models (the small\dsl large pair, i.e.,\ gemma3-12b\dsl gemma4-31b) are served locally with vLLM (v0.20.1) behind OpenAI-compatible endpoints across the 4$\times$ L40S GPUs.
The harness discovery agent (proposer) and \sys's \abstractor{} and \compiler{} run on a frontier model (\texttt{claude-opus-4-7}), while the accuracy judge runs on \texttt{gpt-5-mini}.
These are API calls billed at list rates, and their latency and dollar cost are reported separately from the local open-weight inference.

\smallskip \noindent \textbf{Datasets.}
We use four datasets spanning two functional domains: \emph{retrieval} (finance, legal) and \emph{reasoning} (math, polyreason).
Every dataset uses the same split: (1) search, which the harness discovery agent iterates against (100 queries); (2) in-distribution, which is held out from the same source mix as search and used to select the best harness programs from the Pareto frontier (100 queries); and (3) cross-distribution, which is aligned with search but structurally distinct in document register, question template, or content domain (300 queries).
\cref{tab:datasets} summarizes the datasets.
Note that \emph{domain} refers to the workload we construct, not the benchmark's native form.
For the retrieval domains, we discard each question's gold evidence attachment and point the harness at a pool directory of source documents, so locating the evidence is part of the task.

\smallskip \noindent \textbf{Metrics.}
We report two measures per harness: task \emph{accuracy} (\texttt{gpt-5-mini} judge, per-dataset grading rules) and harness \emph{cost}, the per-question inference cost (\textcent/q).
Accuracy alone is misleading (e.g., a harness can often raise it with more compute), so we report both and compare them together rather than combining into one score~\cite{wei2026moar}; related work likewise treats cost reduction as constrained by an accuracy guarantee rather than traded against it freely~\cite{zeighami2025bargin}.
Results vary from run to run, so we use as many seeds and replicates as our budget allows.

\smallskip \noindent \textbf{Harness Search and Discovery.}
We discover harness programs with Meta-Harness~\cite{metaharness}, a state-of-the-art coding-agent search that browses all prior candidates' source, scores, and traces.
Each search runs $10$ iterations, and we select programs on the Pareto frontier of the accuracy and cost trade-off (in-distribution split)\footnote{\url{https://github.com/stanford-iris-lab/meta-harness}}.
Since \sys's recovery is agnostic to how a harness was produced, we fix the searcher rather than compare searchers.
Every agent also runs inside a sandbox that prevents evaluation leakage.

\begin{table*}[h!]
\caption{Round-trip fidelity and cost, per domain over $3$ seeds $\times\,3$ compile replicates ($=9$ runs); $\Delta=\text{RT}-\text{orig}$. \emph{Compile} is one recover$\rightarrow$compile call; \emph{Meta-Harness search} splits the full search into its proposer and evaluation halves.}
  \label{tab:roundtrip}
  \begin{tabular*}{\textwidth}{@{\extracolsep{\fill}}l ccc ccc cc cccc@{}}
    \toprule
    & \multicolumn{3}{c}{Round-trip accuracy}
    & \multicolumn{3}{c}{Harness \textcent/q}
    & \multicolumn{2}{c}{Compile}
    & \multicolumn{4}{c}{Meta-Harness search} \\
    \cmidrule(lr){2-4}\cmidrule(lr){5-7}\cmidrule(lr){8-9}
    \cmidrule(lr){10-13}
    Domain & Orig & RT & \shortstack{mean\\$\Delta$}
           & Orig & RT & \shortstack{mean\\$\Delta$}
           & \$ & s
           & \shortstack{Prop\\\$} & \shortstack{Eval\\\$}
           & \shortstack{Prop\\wall} & \shortstack{Total\\wall} \\
    \midrule
    Finance    & 0.293 & 0.304 & $+0.011$ & 0.171 & 0.170 & $-0.000$
               & 2.79 & 134 & 67 & 1.4 & 41\,m & 2.8\,h \\
    Legal      & 0.673 & 0.667 & $-0.007$ & 0.420 & 0.413 & $-0.007$
               & 2.80 & 173 & 72 & 2.6 & 41\,m & 4.4\,h \\
    Math       & 0.618 & 0.608 & $-0.010$ & 0.363 & 0.376 & $+0.013$
               & 2.25 & \phantom{0}94 & 44 & 4.5 & 30\,m & 6.8\,h \\
    Polyreason & 0.710 & 0.717 & $+0.007$ & 0.130 & 0.119 & $-0.011$
               & 2.42 & \phantom{0}98 & 43 & 0.8 & 30\,m & 9.2\,h \\
    \bottomrule
  \end{tabular*}
\end{table*}

\subsection{Catalog Construction}
\label{sec:harness-discovery}

Since \sys takes a searched harness as input, we first run Meta-Harness on each domain to produce a strong harness.
Meta-Harness begins its search from a \emph{brief.md} that specifies the task environment and answer interface without supplying a solution strategy.
The search agent can invoke both the small and large models, which permits candidate programs such as small-to-large cascades, and a subagent retains the candidates on the accuracy-cost Pareto frontier. 
As shown in \cref{fig:mh-trajectory}, the three searches for each content domain progress from the brief-only program (i.e., \emph{iter 0}) to the source harnesses selected on the in-distribution split.


\cref{tab:primitive-levels-by-domain} summarizes the primitives recovered from each harness.
Note that retrieval programs are roughly twice the size of reasoning ones ($\approx 26$ vs.\ $14$), carrying extra machinery to index and ground a document pool.
More importantly, every skeleton step is functional; that is, the source-specific step count is $0.0$ across all four domains, confirming that the logical plans cleanly transfer and source-specific details are confined to the decision logic on top.

\subsection{Declarative Fidelity}
\label{sec:sem-fidelity}

Round-trip compilation tests whether the declarative description, which drops the original source code and retains only \sys's typed primitives (skeleton steps, beliefs, policies, prompt templates), can correctly reproduce the behavior of the original source harness.
Choices such as evidence filtering, prompt order, fallback conditions, and sample count can affect behavior even when two programs share the same high-level strategy.
Furthermore, the \compiler{} can draw on its own model knowledge, so matching accuracy by itself would not distinguish faithful reconstruction from an entirely different harness.
We therefore attempt to recover each original source harness by extracting \sys primitives and directly compiling them back, running both harnesses on the same $100$-question in-distribution split.
An accurate recreation of the original harness should match both accuracy and cost, suggesting that our compiled harness faithfully matches the original searched harness rather than relying on the model to solve the task from its general priors.

We run this round-trip (RT) test on the three source harnesses per domain from \cref{sec:harness-discovery}.
We recover each source with the \abstractor{}, compile it back with the \compiler{}, and evaluate both on the same $100$-question in-distribution split, reporting accuracy, per-question cost, and the mean round-trip change $\Delta=\text{RT}-\text{orig}$. We also record the one-time cost and latency of a single recover-and-compile call against the search that produced the original harness.

\begin{table}[t]
  \caption{Catalog composition by transfer scope. Cells are mean ($N{=}3$) counts per program, as step\dsl belief\dsl policy\dsl prompt template.}
  \label{tab:primitive-levels-by-domain}
  \begin{tabular}{@{}l cc c@{}}
    \toprule
    Domain & \shortstack{L1\\{(functional)}}
           & \shortstack{L2\\{(source-specific)}}
           & Total \\
    \midrule
    Finance    & 5.0/6.7/7.3/1.0 & 0.0/2.0/5.3/0.0 & 27.3 \\
    Legal      & 3.3/7.6/7.7/0.0 & 0.0/3.7/2.7/1.0 & 26.0 \\
    Math       & 3.0/4.3/5.3/1.7 & 0.0/0.0/0.0/0.0 & 14.4 \\
    Polyreason & 2.3/3.3/4.3/2.0 & 0.0/1.0/1.3/0.0 & 14.3 \\
    \bottomrule
  \end{tabular}
\end{table}

\cref{tab:roundtrip} shows that \sys's RT preserves harness behavior across all four domains.
We observe no major deviation in either accuracy, with at most a $\pm 0.011$ difference (finance $+0.011$, legal $-0.007$, math $-0.010$, polyreason $+0.007$), or per-question cost (mean $\Delta$ within $\pm0.013$\,\textcent/q).
One RT costs about \$2.3--2.8 and takes only a few minutes, compared to \$43--72 and hours for the original Meta-Harness search.

\begin{figure*}[t]
  \centering
  \includegraphics[width=\textwidth]{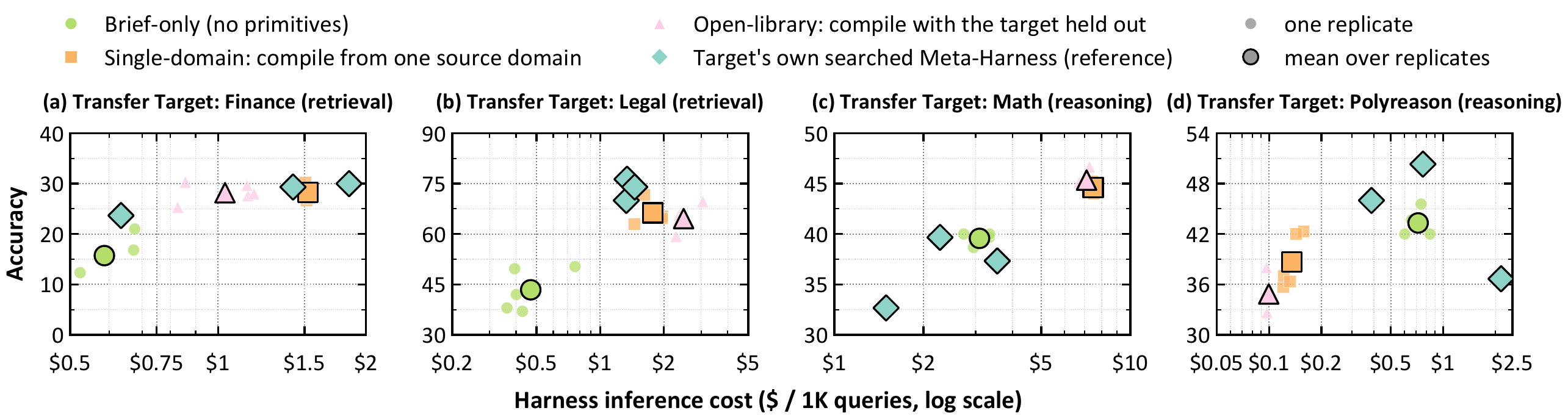}
  \caption{Primitive reuse, one plot per transfer target. Accuracy on the target's cross-distribution split against harness inference cost.}
  \label{fig:crossdomain-transfer}
\end{figure*}

\subsection{Primitive Reuse}
\label{sec:sem-composition}

To determine whether our primitives generalize and can be effectively reused, we compiled harnesses for unseen content domains using functional primitives extracted in the same task family.
We test four source\dto{}target transfers covering retrieval (finance\dlr{}legal) and reasoning (math\dlr{}polyreason).
Every version receives the target brief (i.e., the dataset description and answer interface) and uses the same model pair, cross-distribution split, and judge.
For the compiler, \emph{brief-only} gets no catalog entries, \emph{single-domain} receives the L1 primitives from one specified source, and \emph{open-library} can access the full catalog containing every domain except the target.
The results are shown in \cref{fig:crossdomain-transfer}, with the target's own Meta-Harness harness serving as the reference point.

For retrieval, \sys's primitives improve both target domains.
Legal\dto{}finance (\cref{fig:crossdomain-transfer}a) increases accuracy from $15.7$ to $28.3$, which falls within the run-to-run variation of the searched harness at $27.7$, and finance\dto{}legal (\cref{fig:crossdomain-transfer}b) increases accuracy from $43.4$ to $66.3$, recovering $76\%$ of the distance to the searched result of $73.4$.
The open-library compiler produces similar values ($28.2$ vs. $28.3$ for finance and $64.6$ vs. $66.3$ for legal) even though it must choose among primitives from all domains.
In every replicate, the compiler selects the retrieval and model-routing primitives, rejects those associated with reasoning, and specializes the legal harness through appropriate slots (e.g., contract-attorney persona, clause\dsl yes-no\dsl no-clause output format) without modifying the templates.

Transfer for reasoning depends heavily on the operating point encoded by the source policies.
Applying math primitives to polyreason (\cref{fig:crossdomain-transfer}d) preserves a low inference cost of $\$0.03$--$0.04$ per question ($5$--$7\times$ below brief-only), yet the two compiled variants reach accuracies of $38.7$ and $34.9$, below the brief-only value of $43.3$.
In the opposite direction (\cref{fig:crossdomain-transfer}c), the polyreason harness supplies an ensemble whose adjudication policy uses the format-independent \texttt{answers\_agree} belief, improving math accuracy from $39.6$ to $44.6$ and $45.4$.
Thus, the compiler successfully transfers the recovered structure in both cases, but the selected beliefs and policies determine whether that structure occupies a useful region of the target cost-accuracy frontier, which is something that a future cost-based optimizer might allow the user to directly trade off (see \cref{sec:challenges}).

Lastly, we note that the searched harnesses tend to overfit to the task domain compared to those compiled by \sys.
For example, the searched polyreason harness falls from $72\%$ in-distribution to $47\%$ on the cross-distribution split, and the searched math harness falls from $67\%$ to $37\%$, below the corresponding baseline.
\sys's compiled harnesses do not exhibit these same accuracy drops, even beating the searched program by roughly eight points on polyreason\dto{}math.
Our approach also wins on cost, where one recover-and-compile pass costs about \$2.5 and takes two minutes compared to tens of dollars and hours for per-domain search.

%% file: 4_challenges.tex
\section{Open Challenges}
\label{sec:challenges}

Our preliminary results demonstrate that \sys can effectively recover and reuse declarative primitives from harnesses, but they also expose several open challenges that must be addressed to build a full-fledged system for managing agentic execution.
In the following, we highlight some of these research opportunities.

\smallskip \noindent \textbf{Cost-Based Compilation.}
The current \compiler{} selects primitives using role matching and natural-language metadata, making it a rule-based rewriter rather than a cost-based optimizer.
Since each evaluated harness provides various measurements (e.g., accuracy, per-question inference cost, latency, belief selectivity), a cost-based compiler could use these statistics to enumerate compatible primitive combinations and maximize expected accuracy subject to a cost or latency budget, similar to a traditional query optimizer.
The central challenges are estimating conditional operator quality, accounting for interactions among primitives, and remaining robust when sparse observations produce inaccurate estimates.


\smallskip \noindent \textbf{Declarative Catalog Maintenance.}
A catalog that only grows will eventually degrade by accumulating primitives that are either redundant, stale, erroneous, or dominated by some better version.
Following physical design advisors~\cite{chaudhuri1997autoadmin}, \sys could evict primitives whose accuracy delta falls below a threshold, consolidate similar primitives into generalized specifications with wider slots, and flag primitives whose belief distributions have drifted.
Selectivity drift is a natural staleness signal.
For example, if \texttt{is\_semantic\_query} fired $70\%$ of the time on a retrieval workload and now fires only $20\%$, the policies that depend on it need re-tuning.

\smallskip \noindent \textbf{Execution Context Recovery.}
\sys currently recovers the harness program, but execution also depends on other context (e.g., system prompts, skill definitions, tool configurations) that may also result from search.
As an example, the remaining finance\dto{}legal accuracy gap from \cref{sec:sem-composition} may reflect context that \sys's current \abstractor{} does not capture.
Extending recovery to these artifacts would allow the same transfer-scope and provenance machinery to distinguish potentially functional context (e.g., a retrieval persona or chunking strategy) from task-specific settings.

\smallskip \noindent \textbf{Pluggable Backends.}
The catalog stores substrate-independent optimization knowledge, so a recovered description could target execution systems other than Python harness code similar to how a logical query plan is decoupled from physical operator implementations.
A LOTUS~\cite{patel2025lotus} backend, for example, could lower recovered steps into semantic operators with per-operator accuracy guarantees.
Similarly, a backend that produces formal specifications could expose the assumptions required to close the gap between \sys's current behavioral fidelity tests and the semantic-equivalence guarantees available in the traditional UDF setting~\cite{ramachandra2017froid,duta2020plsql}.
Both directions require a precise contract for probabilistic or LLM-defined operators whose behavior can change across models and prompts.

%% file: 5_related.tex
\section{Related Work}
\label{sec:background}

\sys lies between systems that construct LLM workflows and database techniques that recover and optimize declarative structure.
We compare these areas by the artifact available after construction, specifically whether it exposes the computation, runtime information, and resulting decisions independently of the original program.

Most orchestration frameworks retain the harness in the form in which a developer authored it.
LangChain~\cite{langchain} and AutoGen~\cite{wu2023autogen}, for example, represent control flow through program structure, prompt text, and inter-agent conversation, whereas agentic prompting methods such as ReAct~\cite{yao2023react} delegate more of the control flow to model reasoning.
Declarative LLM data systems provide operators for specifying a computation and may optimize aspects of its execution~\cite{patel2025lotus,liu2025palimpzest,shankar2025docetl,khattab2024dspy}.
The recovery problem studied in this work begins after another process has already constructed the workflow.

Coding agents can search over complete harness programs by examining the source, evaluation results, and traces produced by earlier candidates~\cite{metaharness}, applying the evolutionary program search that coding agents use for algorithms and code more broadly~\cite{novikov2025alphaevolve,sharma2025openevolve}.
Prompt optimizers similarly search a narrower space and retain an optimized textual artifact for a fixed program~\cite{agrawal2026gepa}, as do context optimizers that accumulate structured playbooks alongside it~\cite{zhang2026ace}.
Program-level search can discover changes to operator order, conditional execution, model selection, and prompting in a single run, but its output combines these decisions in one implementation.
\sys accepts these results as input and attempts to recover reusable primitives at a granularity that permits separate cataloguing and transfer.

As mentioned, procedural UDF recovery supplies a database precedent for this separation.
Systems such as Froid expose expressions and dependencies from imperative functions so that an optimizer can reason across the former procedural boundary~\cite{ramachandra2017froid,duta2020plsql}, and related work synthesizes queries from imperative database-backed application code~\cite{cheung2013qbs}.
\sys adopts the representation-level objective, although its operators may invoke stochastic models and therefore lack a comparable algebra of semantics-preserving rewrites.
Automatic tuning provides a complementary precedent in which physical design advisors choose access structures from workload cost estimates~\cite{chaudhuri1997autoadmin}, potentially even learning from prior measurements to select DBMS settings~\cite{vanaken2017ottertune,zhang2019cdbtune}; more recent work applies LLM-driven optimization to systems design beyond the database itself~\cite{hamadanian2025glia}.
Harness compilation introduces different quantities to estimate, including output quality, model inference cost, and the selectivity of derived runtime attributes.

Some prior work attempts to reuse intermediates produced by agentic search.
Existing approaches include structural priors over typed operators~\cite{du2026swift}, latent capability bases~\cite{wang2026capflow}, code-level abstraction operators~\cite{zhao2026a2flow}, and repositories of reusable workflows~\cite{yuan2026flowbank,wang2024awm}.
Skill-library methods distill programs into reusable functions~\cite{stengeleskin2024regal} or accumulate them from successful trajectories~\cite{wang2023voyager}.
These artifacts range from individual capabilities to complete programs, whereas \sys focuses on the intermediate plan representation needed to manage transfer as a systems decision.

%% file: 6_conclusion.tex
\section{Conclusion}
\label{sec:conclusion}

This paper made the case for treating searched harnesses as managed declarative artifacts rather than retaining only their imperative implementations.
\sys recovers a typed description of a harness, records its primitives and provenance in a catalog, and compiles compatible structure into harnesses for new domains without another search.
The preliminary evaluation shows that round-trip compilation closely preserves measured accuracy and cost while also demonstrating effective transferability across domains.

Beyond harness search, agentic workload execution also produces significant optimization knowledge that is currently discarded.
The research agenda we described lays out a path toward making this information persistent and reusable, and \sys is a first step in that direction.
These are database systems problems (e.g., cost-based optimization, declarative specification of primitives), and the database community is well-positioned to solve them.
